\documentclass[journal]{IEEEtran}
\ifCLASSINFOpdf
\else
\fi
\usepackage{amsmath}
\usepackage{epsfig}
\usepackage{graphicx}
\usepackage{amssymb}
\usepackage{multirow}
\usepackage{booktabs}
\usepackage{amssymb}
\usepackage{color}
\usepackage{url}
\usepackage{cite}
\usepackage{hyperref}

\begin{document}
%
\title{Towards Effective Multi-Moving-Camera Tracking: A New Dataset and Lightweight Link Model}
%
%
%

\author{Yanting Zhang,~\IEEEmembership{Member,~IEEE,}
        Shuanghong Wang,~\IEEEmembership{Graduate Student Member,~IEEE,}
        ~Qingxiang~Wang, \\Cairong Yan,~\IEEEmembership{Member,~IEEE,} and Rui Fan,~\IEEEmembership{Senior Member,~IEEE} 

\thanks{This work is supported by National Natural Science Foundation of China (62206046) and Shanghai Sailing Program (21YF1401300). 
}
\thanks{Yanting Zhang, Shuanghong Wang, and Cairong Yan are with the School
of Computer Science and Technology, Donghua University, Shanghai 201620, P. R. China. Qingxiang Wang is with the Institute of Integrated Science and Technology, Hosei University, Tokyo  1848584, Japan. Rui Fan is with the Machine Intelligence \& Autonomous Systems (MIAS) Group, the Robotics \& Artificial Intelligence Laboratory (RAIL), the College of Electronic \& Information Engineering, the State Key Laboratory of Intelligent Autonomous Systems, and Frontiers Science Center for Intelligent Autonomous Systems, Tongji University, Shanghai 201804, P. R. China. 
}


}

%
%

\markboth{IEEE Transactions on Intelligent Vehicles}%
{Shell \MakeLowercase{\textit{et al.}}:Bare Demo of IEEEtran.cls for IEEE Journals}
%



\maketitle

\begin{abstract}
Ensuring driving safety for autonomous vehicles has become increasingly crucial, highlighting the need for systematic tracking of on-road pedestrians. 
Most vehicles are equipped with visual sensors, however, the large-scale visual data has not been well studied yet. 
Multi-target multi-camera (MTMC) tracking systems are composed of two modules: single-camera tracking (SCT) and inter-camera tracking (ICT). To reliably coordinate between them, MTMC tracking has been a very complicated task, while tracking across multiple moving cameras makes it even more challenging. In this paper, we focus on multi-target multi-moving-camera (MTMMC) tracking, which is attracting increasing attention from the research community. Observing there are few datasets for MTMMC tracking, we collect a new dataset, called Multi-Moving-Camera Track (MMCT), which contains sequences under various driving scenarios. To address the common problems of identity switch easily faced by most existing SCT trackers, especially for moving cameras due to ego-motion between the camera and targets, a lightweight appearance-free global link model, called Linker, is proposed to mitigate the identity switch by associating two disjoint tracklets of the same target into a complete trajectory within the same camera. Incorporated with Linker, existing SCT trackers generally obtain a significant improvement. Moreover, to alleviate the impact of the image style variations caused by different cameras, a color transfer module is effectively incorporated to extract cross-camera consistent appearance features for pedestrian association across moving cameras for ICT, resulting in a much improved MTMMC tracking system, which can constitute a step further towards coordinated mining of multiple moving cameras. The project page is available at \href{https://dhu-mmct.github.io/}{https://dhu-mmct.github.io/}.

\end{abstract}


%
\IEEEpeerreviewmaketitle


\section{Introduction}
\label{sec:intro}


Autonomous driving has garnered widespread attention globally in recent times. With the ability to reduce accidents, improve traffic flow, and increase accessibility for individuals who are unable to drive, autonomous vehicles have the power to greatly enhance mobility and safety for all. However, some critical challenges remain in terms of safe driving \cite{yurtsever2020survey}. For example, pedestrians can appear outside of the field of view (FoV) of the camera on a driverless car, but they might be visible in other vehicles’ FoVs. Currently, different visual data collected from distinct moving vehicles are analyzed independently. Without a coordinated mining process, significant inter-connected information among vehicles cannot be obtained. In general, a well-developed perception system can result in better driving status. 

Tracking plays a significant role in autonomous driving. To be specific, a satisfactory performance of pedestrian tracking is conducive to subsequent tasks of autonomous vehicles. Thanks to the Multi-Object Tracking (MOT) Challenge and the released dataset \cite{milan2016mot16}, human tracking under a single moving camera gradually becomes mature \cite{wang2020towards}. However, it is sometimes difficult to track multiple targets within a single camera due to heavy occlusion, which makes the association performance of noisily detected targets further degraded, not to mention the further complication caused by the ego-motion due to moving cameras, resulting in a higher rate of identity (ID) switches. 
Thus, it is important to come up with a global link method to fix such problems. 

Multi-target multi-camera (MTMC) tracking
is a challenging problem since the illumination, viewpoints, and video quality of each camera can vastly vary \cite{amosa2023multi}. The existing methods for MTMC tracking are mostly related to multiple static cameras. There are few publicly available annotated datasets for tracking under multiple moving cameras \cite{zhang2017multi,hou2019locality,he2020multi}. More specifically, Lee \textit{et al.} \cite{lee2014driving} develop a Visual Simultaneous Localization and Mapping (V-SLAM) system to collaboratively reconstruct the 3-D objects from multi-view videos. However, the tracking issues across moving cameras are not considered. Zhang \textit{et al.} \cite{zhang2021pedestrian} propose a multi-target multi-moving-camera (MTMMC) dataset, called ``DHU-MTMMC”, and explores MTMMC tracking tasks taking advantage of some deep learning approaches. However, a more effective SCT method is expected to obtain better results, and a larger dataset is needed for a more convincing evaluation. Zhang \textit{et al.} \cite{zhang2022road} utilize FairMOT \cite{zhang2021fairmot} to extract discriminative appearance features for both single-camera and multi-camera data associations. However, the image style variations caused by cameras are not considered.

In this paper, an MTMMC tracking pipeline is established to coordinately mine the visual data from multiple moving cameras. To further justify the proposed MTMMC tracking workflow with improved experimental results, we conduct extensive experiments involving global link methods and a color transfer function for discriminative appearance features on the proposed ``MMCT” dataset. Overall, the contributions of this paper are summarized as follows:
\begin{itemize}

\item We propose a new ``MMCT” benchmark dataset, which is collected from several cameras mounted on moving cars. It characterizes longer sequences and higher density for evaluating trackers aiming at tracking under multiple moving cameras. 

\item  We propose a systematic tracking workflow to perform multi-moving-camera pedestrian tracking. In particular, a novel appearance-free global link model, Linker, is proposed to address the common missing association problems in SCT, especially for moving camera scenarios. It can be integrated with any existing SCT tracker to improve its tracking results at a negligible extra cost. Moreover, leveraging a color transfer module, robust and cross-camera consistent appearance features are extracted for the association in multi-camera tracking.
\item Extensive experiments on the proposed MMCT and DHU-MTMMC datasets are conducted, which demonstrate the effectiveness of our on-road pedestrian tracking framework and the proposed Linker in resolving association failures. 
\end{itemize}





\section{Related Work}

\subsection{Detection and Embedding}
\subsubsection{Separate detection and embedding}
Most MOT methods follow the paradigm of tracking-by-detection, and thus detection is a fundamental step. Detection quality has a big impact on the final performance of the tracker, and it is always beneficial to adopt a high-accuracy detection algorithm to boost the subsequent steps in tracking. CenterNet \cite{zhou2019objects} utilizes the center point of the object in the image to detect the target, and predicts the coordinates of the object's center point as well as its width and height to locate the object. ByteTrack \cite{zhang2022bytetrack} achieves significant improvement on IDF1 score by associating every detection box instead of only the high score ones.

In previous works, traditional appearance-based embedding features are represented by color \cite{forssen2007maximally}, texture descriptors \cite{lowe2004distinctive, ojala2002multiresolution}, and the patch-based/local descriptors \cite{farenzena2010person}. However, such handcrafted features can not give discriminative high-level representations with semantic meaning. Current mainstream approaches usually adopt deep embedding features obtained through convolutional neural networks (CNNs).
For example, DeepSORT \cite{wojke2017simple} uses a simple CNN which is pretrained on the person re-identification dataset MARS \cite{zheng2016mars} to extract embedded appearance features. In addition to the appearance-based embedding features, motion patterns of bounding boxes are effectively taken into account heuristically during association as in DeepSORT for online tracking. To explore the significance of motion patterns for tracking, Wang \textit{et al.} \cite{wang2021track} propose a reconstruct-to-embed strategy to embed tracklet based on deep graph convolutional neural networks (GCNs), which achieves a competitive performance though without appearance information. Moreover, appearance features can be integrated with motion patterns for subsequent association and offline tracking \cite{wang2019exploit}. Note that cross-camera \cite{lee2015road} settings are considered an even tougher task, where the embedding acts as a vital step.

\subsubsection{Joint detection and embedding}
To save the computation resources better, joint detection and embedding (JDE) based trackers perform object detection and feature embedding, which can be jointly learned, through a shared single network. Wang \textit{et al.} \cite{wang2020towards} propose such a JDE model that can simultaneously output detection and the corresponding embedding features, which reports the first (near) real-time online MOT algorithm design. Zhang \textit{et al.} \cite{zhang2021fairmot} further present FairMOT, which adds an additional embedding branch for CenterNet \cite{zhou2019objects} to extract and output features in the corresponding center of the detected target. Thanks to its effectiveness in tackling unfairness caused by presumed anchors, features, and feature dimensions, FairMOT has obtained high detection and tracking accuracy and greatly outperformed previous state-of-the-arts in SCT. In this paper, our single-camera-tracking approach is based on FairMOT, ensuring good SCT performance.

\subsection{Data Association in Tracking }
\subsubsection{Graph-aided tracking}
With the detections and embedding features being ready, a standard way to perform data association is by using a graph, where each detection serves as a node and edges indicate possible links among them \cite{ristani2018features}. The data association can then be formulated as maximum flow \cite{berclaz2006robust} or, equivalently, minimum cost problem with either fixed costs based on distance \cite{pirsiavash2011globally, zhang2008global} or learned costs \cite{leal2014learning}. Wang \textit{et al.} \cite{wang2019exploit} treat a previously established tracklet as a node and designs a TrackletNet to measure the connectivity between two tracklets. These graph-based methods need to perform computationally expensive global optimization on large graphs, which limits their application to online tracking.

\begin{figure*}[h]
  \centering
  \includegraphics[width=0.85\linewidth]{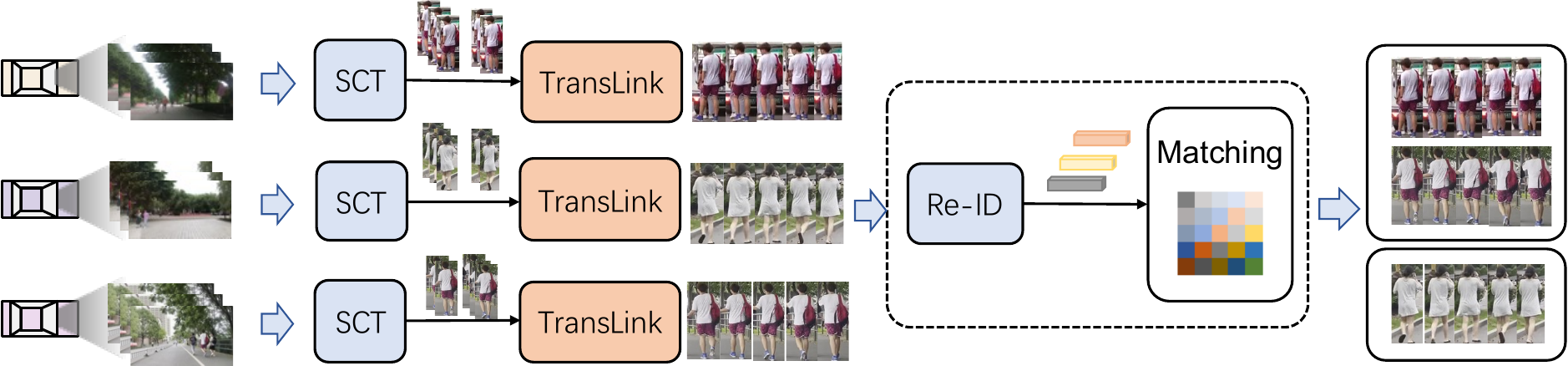}
    \caption{The workflow of our MTMMC tracking, composed of three major parts: single camera tracking (SCT), global link model of Linker for refining SCT results, and cross-camera tracklet matching.}
  \label{MTMC}
\end{figure*}

\subsubsection{Transformer-aided tracking}
Recently, Transformer \cite{vaswani2017attention} has shown great success in various visual tasks, such as image classification \cite{dosovitskiy2020image}, detection \cite{carion2020end}, segmentation \cite{liang2020polytransform}, human pose estimation \cite{zheng20213d}, as well as tracking \cite{meinhardt2021trackformer, sun2020transtrack, zeng2021motr}. Transformer-based frameworks for MOT achieve implicit data association between frames via attention. Both TransTrack \cite{sun2020transtrack} and TrackFormer \cite{meinhardt2021trackformer} involve object queries, which take care of new detections, and track queries, which are responsible for tracking objects across frames. 

\subsubsection{Global link in tracking}
MOT has achieved great improvements in recent years. However, some challenges still exist. For example, identity switches are likely to occur when the object is temporally occluded or the target exhibits deformation. Some powerful global link methods have been proposed to diminish such problems. Yang \textit{et al.} \cite{yang2021remot} take the initiative to split imperfect tracklets and then merges the split tracklets by taking advantage of appearance features. Wang \textit{et al.} \cite{wang2022split} develop a tracklet booster consisting of two parts, \textit{i.e.}, Splitter and Connector, which can effectively tackle association errors. However, the above approaches rely on appearance features, which results in higher sensitivity to appearance variance. StrongSORT \cite{du2022strongsort} proposes a global link model, termed AFLink, which only exploits motion information but can achieve a good performance. TransLink \cite{zhang2023translink} leverages the self-attention mechanism to obtain representative global features by fully exploiting the appearance and motion features, and achieves promising results. Inspired by the aforementioned methods, in this paper, an appearance-free link model, called Linker, is proposed to reduce missing association errors, with fewer parameters but higher efficiency achieved compared to AFLink and TransLink.

\subsection{Inter-Camera Tracking and Style Transfer}
\subsubsection{Inter-Camera Tracking}
MTMC tracking needs to perform tracking across cameras, \textit{i.e.}, inter-camera tracking (ICT), where data association is more difficult than that of SCT, since various cameras placed at different sites usually show distinct lighting conditions, viewing angles, \textit{etc.} Besides, the pedestrians' motion patterns across different cameras are hard to be explicitly expressed. Huang \textit{et al.} \cite{huang1997object} are the first to perform non-overlapping multi-camera tracking utilizing color and spatiotemporal features with a Bayesian tracking model. Qian \textit{et al.} \cite{qian2020electricity} design an aggregation loss for training in the Re-ID part, aiming at eliminating the appearance appearance variance across different FoVs. Shim \textit{et al.} \cite{shim2021multi} implement MTMT tracking following a standard procedure, \textit{i.e.}, detection and feature extraction, single camera tracking, and multi-camera association of trajectories from each camera. Nguyen \cite{nguyen2023multi} presents an MTMC tracking system with the introduction of R-matching algorithm and Gaussian mixture model, which performs remarkably well in both synthetic and real data. Hao \textit{et al.} \cite{hao2023divotrack} provide a novel baseline cross-view multi-object tracking method named CrossMOT, which is the first work that extends the joint detection and embedding from a single-view tracker to the cross-view.


\subsubsection{Style Transfer}
Style transfer is commonly applied in pedestrian re-identification tasks \cite{zhong2018camera, rami2022online}. To mitigate the influence of image style variations caused by different cameras, Zhong et al. \cite{zhong2018camera} propose a method called CamStyle (Camera Style) for style transformation. Rami et al. \cite{rami2022online} introduce a challenging online unsupervised domain adaptation (OUDA) setting for pedestrian re-identification. The style transfer module in pedestrian re-identification tasks typically relies on complex deep learning models. Cui et al. \cite{cui2023novel}, focusing on the geolocation task, use a simple color transfer function to reduce the differences between satellite and drone images, resulting in improved geolocation accuracy. Inspired by this research, this paper applies the color transfer function for cross-camera data association. This method is simple and efficient, significantly reducing the style differences among different camera images.

\section{Proposed MTMMC Tracking Methodology}

In this section, we decompose the workflow of MTMMC tracking into three major parts, as shown in Figure \ref{MTMC}. We first obtain tracklets under each camera utilizing the FairMOT method which simultaneously outputs detection and the corresponding Re-ID features, which are then used for cross-camera data association. Linker can be applied for better SCT results. It's noted that Linker is play-and-plug. Finally, we connect the tracklets across cameras to obtain matching results. Our MTMMC tracking framework provides a general solution for coordination among multiple moving cameras. It can adopt any newer novel and effective SCT methods.

\subsection{Global Link of Tracklets in SCT}
\begin{figure}[h]
  \centering
  \includegraphics[width=0.95\linewidth]{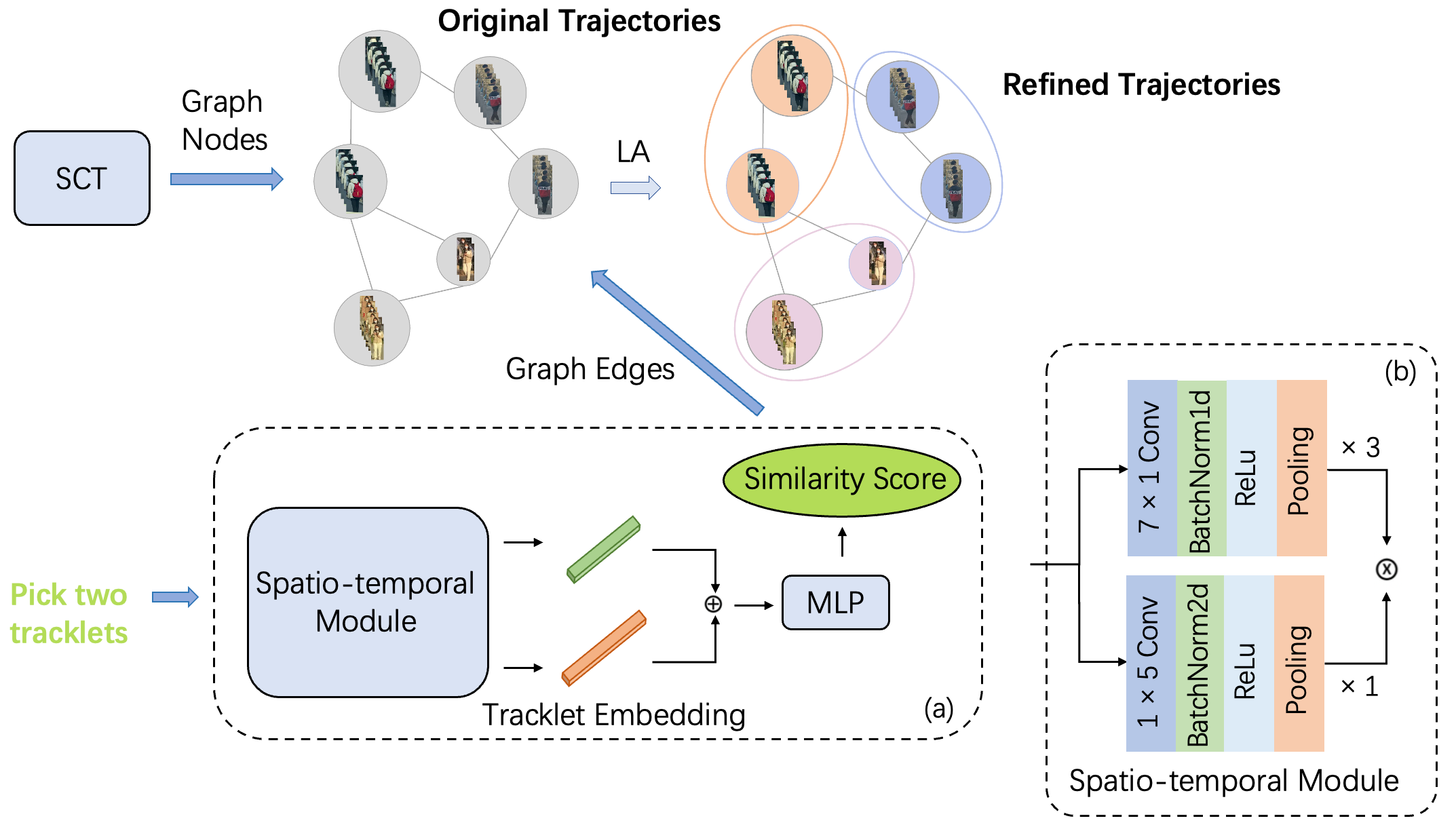}
    \caption{The workflow of the Linker model. LA refers to the linear assignment algorithm. Given the output short tracklets of SCT trackers and the similarity score, a tracklet graph is constructed. After that, A modified Jonker-Volgenant algorithm is leveraged to solve the assignment problem.}
  \label{Linker}
\end{figure}

To address the common identity switches faced by SCT trackers under a moving camera due to ego motions and occlusion when the appearance feature becomes unreliable, a lightweight appearance-free model, called Linker, is proposed to associate two disjoint tracklets of the same target into a complete trajectory. Linker is used after SCT when the short tracklets are collected. To mitigate the impact of camera motion and occlusion, Linker incorporates the motion features of the trajectories as input. A tracklet lasting for $l$ frames is denoted as $\boldsymbol{T} = \left \{ I_{k},x_{k}, y_{k}, w_{k}, h_{k}\right \}_{k=1}^{l}$, where $I$ denotes the frame index and $(x,y,w,h)$ signifies the box’s location and size which represent the motion information of the tracklet. Figure \ref{Linker}(a) exhibits the architecture of the Linker model, whose inputs are two tracklets $\boldsymbol{T}_{i}$ and $\boldsymbol{T}_{j}$ of the recent $N$ frames within their lifelong lengths. For tracklets that are shorter than $N$ frames, they are padded with the starting or ending frames. 

The motion representations of two tracklets are fed into a spatiotemporal module separately to generate the tracklet embeddings. The spatiotemporal information is fully explored through the module, which involves two parallel branches, as shown in figure \ref{Linker}(b). The upper branch handles the temporal correlation by convolving along the temporal dimension with $7\times 1$ kernels. The lower branch tries to integrate the spatial information by applying $1\times 5$ convolutions to fuse information from different feature dimensions, namely $I$, $x$, $y$, $w$, and $h$. Each convolution is then followed by a Batch Normalization (BN) layer \cite{ioffe2015batch}, and a ReLU activation layer \cite{glorot2011deep}. 


After each tracklet is passed through the spatial and temporal branches separately, we fuse the two resulting features of each tracklet with element-wise multiplication. Finally, the two tracklet embeddings are concatenated as the input denoted as $E_o$ of a multilayer perceptron (MLP) to predict an association confidence score, which indicates how likely the two tracklets belong to the same identity. The MLP consists of two fully connected layers with a ReLU activation layer inserted in between. There are two neurons at the last layer to predict the scores for two classes. The output logits $s_{0}$ and $s_{1}$ are passed to the Softmax function to obtain a predicted probability of $\hat{p}$ how likely the two tracklets belong to the same identity. The binary cross-entropy loss $L$ is used in the training as follows,


\begin{equation}
L = -p\log \left ( \hat{p} \right )-\left ( 1-p \right )\log \left ( 1- \hat{p} \right ),
\end{equation}
where $p$ denotes the ground truth label.

Based on the probability $\hat{p}$ that two tracklets belong to the same identity and the output short tracklets of SCT trackers, a tracklet graph is constructed with each tracklet as a vertex and pairs of two tracklets as the edge, as shown in Figure \ref{Linker}. The weight of each edge is set as $1-\hat{p}$ representing the connectivity cost of $\boldsymbol{T}_{1}$ and $\boldsymbol{T}_{2}$. Tracklet pairs, with confidence scores lower than threshold $\sigma _{a}$, are neglected for the latter association. A modified Jonker-Volgenant algorithm \cite{crouse2016implementing} is utilized to solve the linear assignment problem. To effectively generate tracklet pairs and reject unreasonable ones, we only consider two tracklets that are not too far from each other, \textit{i.e.}, we set a tolerant time interval between two tracklets as $\sigma _{t}$ and a tolerant spatial difference as $\sigma _{s}$. 

\subsection{Multi-Camera Tracking}

Re-ID features are obtained for each trajectory during the single-camera tracking process using the FairMOT method. Specifically, the 128-dimension Re-ID feature vectors are extracted from the target centers. Besides as a key gradient for data associations between frames in SCT, the extracted Re-ID features can also be employed in cross-camera tracklet association for efficiency and convenience. A pedestrian's tracklet appearance feature is represented by the mean feature vector over multiple frames to relieve the influences from variations in pedestrians’ poses or environmental illuminations over time. The mean feature vector is calculated using the latest 30 frames of the tracklet from one moving camera, and then it is used to associate other tracked pedestrians by comparing it to the mean feature vectors of all the tracklets in other cameras. 


\begin{figure}[h]
  \centering  \includegraphics[width=\linewidth]{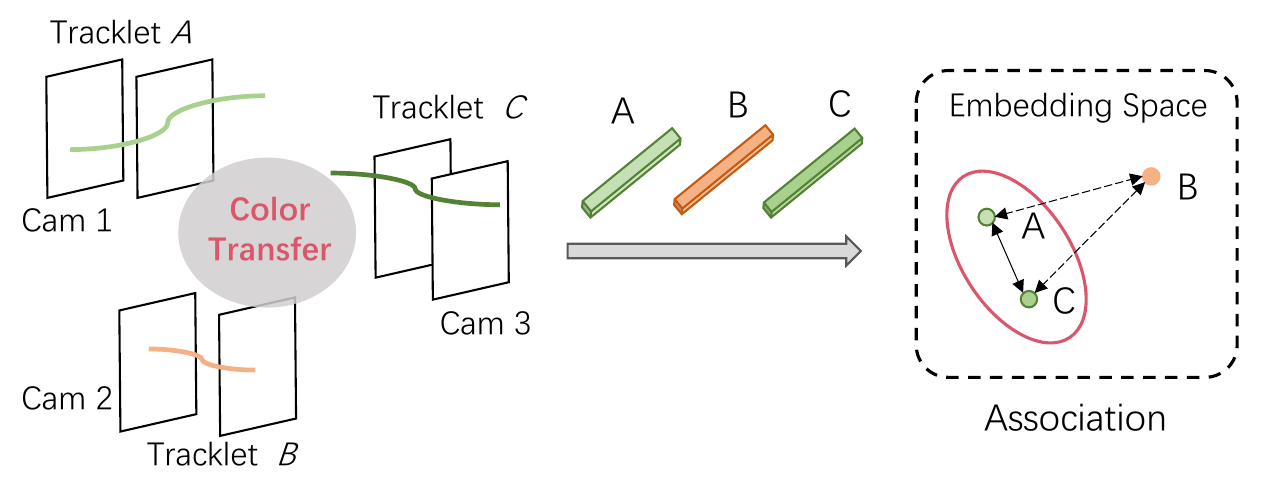}
    \caption{Tracklet embedding association in multi-camera tracking. A color transfer module is employed to extract cross-camera consistent features.}
  \label{Tracklet_association}
\end{figure}

To be specific, the mean feature vectors of the detected persons in one camera are separately compared to those in other cameras. We use cosine distance between feature vectors to determine the dissimilarity of the feature vectors. As is presented in Figure \ref{Tracklet_association}, tracklet pairs that are close in embedding space tend to be associated with each other. We denote $\boldsymbol{F}$ and $\boldsymbol{Q}$ as the tracklet feature matrices in two cameras. $N$ and $M$ are the numbers of tracklets of $\boldsymbol{F}$ and $\boldsymbol{Q}$. The distance matrix $\boldsymbol{D}_{M\times N}$ is computed as follows:




\begin{equation}
\boldsymbol{D} = d(\boldsymbol{Q}, \boldsymbol{F}),
\label{D}
\end{equation}
where $d(\cdot,\cdot)$ stands for the distance function. Each value in $\boldsymbol{D}$ is defined by the cosine distance between two tracklets’ feature vectors. We use threshold $\alpha$ as the association threshold to suppress the mismatches, \textit{i.e.}, and distance values over $\alpha$ in $\boldsymbol{D}$ are not considered in the later association step. Jonker-Volgenant algorithm \cite{jonker1987shortest} is used to find the potential matches for the multi-camera tracking purpose based on the obtained matrix $\boldsymbol{D}$. 

In general, there are significant style differences in images captured by different cameras in the same scene, which poses challenges for cross-camera data association. The color-based style transfer method is employed to reduce the style disparities among images captured by different cameras. To be specific, the color-based style transfer technique is employed to transfer the color characteristics of images captured by different cameras. This process generates color-transferred images, which are subsequently utilized as inputs for the SCT module. By leveraging the color-transferred images, the SCT module extracts appearance features that are more consistent across different cameras, thereby facilitating cross-camera association.

The original images of different scenes are stored in the RGB color space. The RGB color space is based on three primary colors: red, green, and blue, which can be combined to produce a wide range of colors. However, the RGB color space is non-orthogonal, and the values in the R, G, and B channels are not independent. Therefore, assuming that the three channels follow a normal distribution, they cannot be independently transmitted based on the mean and standard deviation of each channel. Hence, for each image, the RGB color space can be transformed into the orthogonal $l \alpha \beta $ color space.

Orthogonal linear transformation is used to transfer the red, green, and blue signals perceived by the human visual system into three uncorrelated color components. It consists of a luminance component $l$ and two chrominance components $\alpha$ and $\beta$. The images captured by Device 1 are considered as the source image $S^{i}$, and the images from other cameras are considered as target images $T^{i}$, where $i=\left \{ l, \alpha, \beta \right \}$. Then, the standard deviation $\sigma _{s}^{i}$ and mean $\hat{S}^{i}$ of the image by Device 1 are computed for each channel in the $l \alpha \beta $ color space, along with the standard deviation $\sigma _{t}^{i}$ and mean $\hat{T}^{i}$ of the target image. The ratio of standard deviations between the source and target images determines the magnitude of color transformation. Therefore, for the target image $M^{i}$, the color transformation formula is as follows:

\begin{equation}
M^{i} = \frac{\sigma _{t}^{i}}{\sigma _{s}^{i}}\left ( S^{i}-\hat{S}^{i} \right )+\hat{T}^{i}.
\label{D}
\end{equation}



\section{Dataset}
In order to justify the proposed MTMMC tracking algorithm and evaluate the performance, we propose an MTMMC dataset, \textit{i.e.}, MMCT, which is a larger dataset than DHU-MTMMC. Furthermore, to better train some of the modules used in this framework, we also use several publicly available benchmark SCT and MTMC tracking datasets for training purposes, such as MOT17 and DukeMTMC-reID datasets.
\subsection{Proposed MMCT dataset}
To the best of our knowledge, there are few public datasets for pedestrian tracking under multiple moving camera recorders. Zhang \textit{et al.} \cite{zhang2021pedestrian} propose the DHU-MTMMC dataset, which will be described in the next subsection, aiming at pedestrian tracking across multiple moving cameras. However, the lengths of some sequences are too short, the density of several sequences is relatively low and the camera motion is smooth.  To compensate for these shortcomings, we further introduce the new MTMMC dataset called MMCT, a set of 32 sequences with more crowded scenarios, longer sequence lengths, more viewpoint variances, and shakier camera motions. The weather condition in MMCT is cloudy, making it a more challenging dataset compared to DHU-MTMMC with sunny weather conditions. These two datasets are then complementary to each other for better evaluating the MTMMC tracking algorithms. 

\begin{figure}[h]
  \centering
  \includegraphics[width=\linewidth]{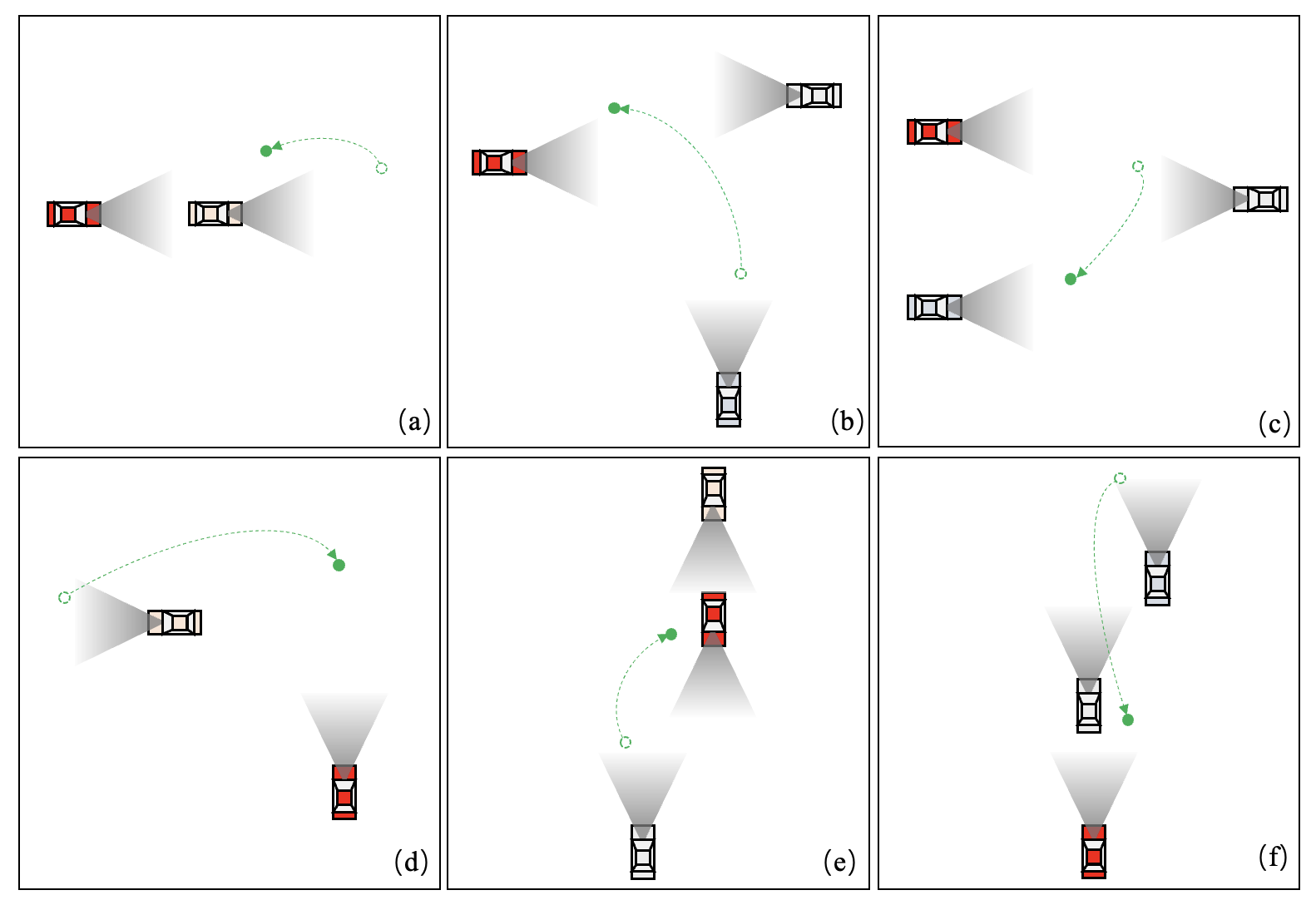}
    \caption{Different driving cases considered during the data collection. Some possible exampled pedestrian movements in green color are also shown.}
    \label{capturedata}
\end{figure}

We have utilized mobile phones placed on three cars to collect the video data. The ``SensorLog" software is pre-installed on mobile devices to record the Global Positioning System (GPS) information. The configurations of different phones are shown in Table \ref{Configurations of the devices}. Note that, the videos are recorded at 30 frames per second (FPS), though we only sample and annotate 5 frames per second to create the dataset. These cars are driven within a campus to record the videos. To make scenarios more complicated, we take different driving cases into consideration, as shown in Figure \ref{capturedata}.

\begin{table}[]
\caption{Configurations of the devices used in video collection.}
\centering
\begin{tabular}{cccc}
\hline
Device & Type           & Resolution         & FPS \\ \hline
1      & Iphone 12 Mini      & 1920 $\times$ 1080 & 30  \\
2      & Iphone 6S & 1920 $\times$ 1080 & 30  \\
3      & Iphone 8       & 1920 $\times$ 1080 & 30  \\
4      & Android Redmi  & 1920 $\times$ 1080 & 30  \\ \hline
\end{tabular}
\label{Configurations of the devices}
\end{table}

\begin{figure}[h]
  \centering
  \includegraphics[width=\linewidth]{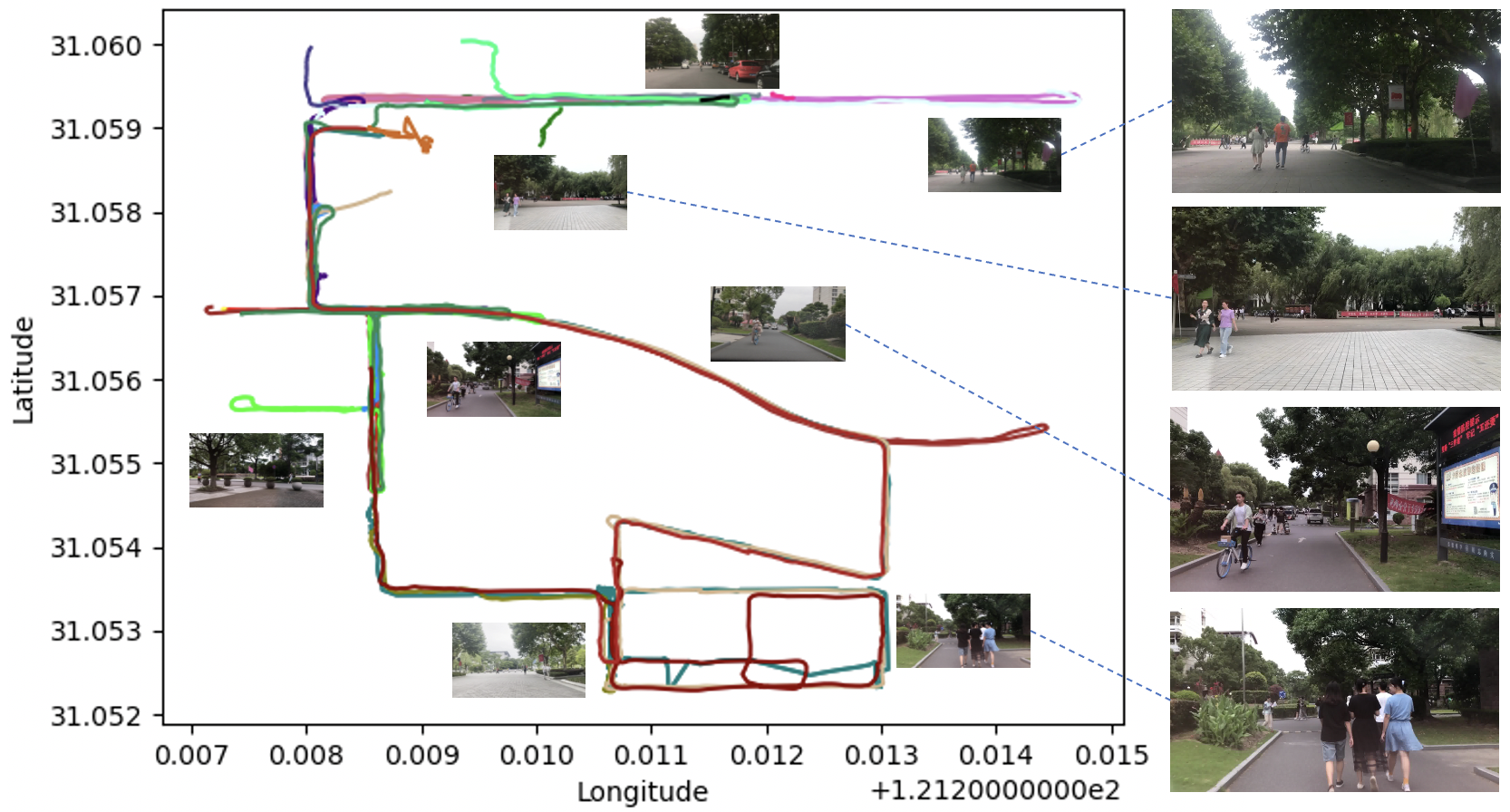}
    \caption{Overview of the MMCT dataset. Each colored trajectory on the left represents a driving segment from a moving camera. Four images on the right show exemplary road scenes.}
    \label{overview_dataset}
\end{figure}

Figure \ref{overview_dataset} exhibits an overview of the new proposed dataset, MMCT, focusing on multi-target multi-moving-camera tracking. The trajectories in Figure \ref{overview_dataset} are plotted based on the detailed GPS cues. As is shown in Figure \ref{overview_dataset}, the routes are complicated, which is conducive to observing a large number of pedestrians in different scenarios. The videos captured from different devices can be synchronized utilizing global timestamps. We select 12 distinct scenarios from all the recordings to ensure the diversity of the dataset. Each scenario may include the interaction of two or three cameras on different cars. 

We manually annotated the pedestrians in the video using annotation tools, following the format of the MOT dataset. The annotations include bounding box information and corresponding trajectory ID. For partially occluded pedestrians, we annotate the complete bounding boxes. For completely occluded pedestrians, we do not provide annotations. When a pedestrian reappears in subsequent frames, its identity ID remains consistent with the ID assigned when it is visible in the previous frame. To facilitate the annotation process, we first utilize an object detector to obtain preliminary bounding boxes. Our annotation tools automatically propagated the bounding boxes from previous frames to the current frame, allowing annotators to make minor adjustments to the boxes in the current frame. To improve the quality of annotations, we conduct repeated checks on the annotation results, making necessary adjustments to rectify any errors.

\begin{figure}[h]
  \centering
  \includegraphics[width=\linewidth]{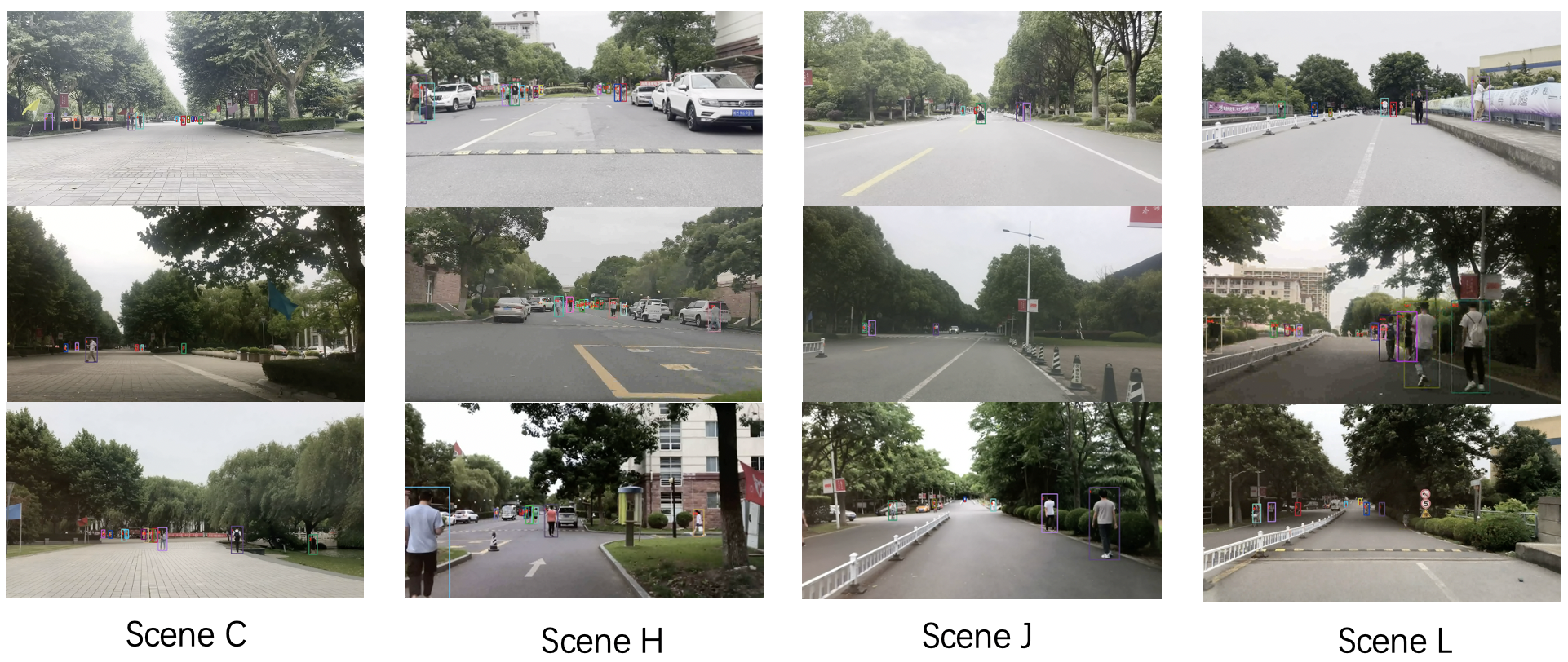}
    \caption{Exemplary video frames of sequences I, II, and III in Scene C, H, J, and L. The color brightness in different sequences captured by various devices is slightly different. The tracked boxes are also drawn in the images.}
  \label{data_scene}
\end{figure}

\begin{figure}[h]
  \centering
  \includegraphics[width=\linewidth]{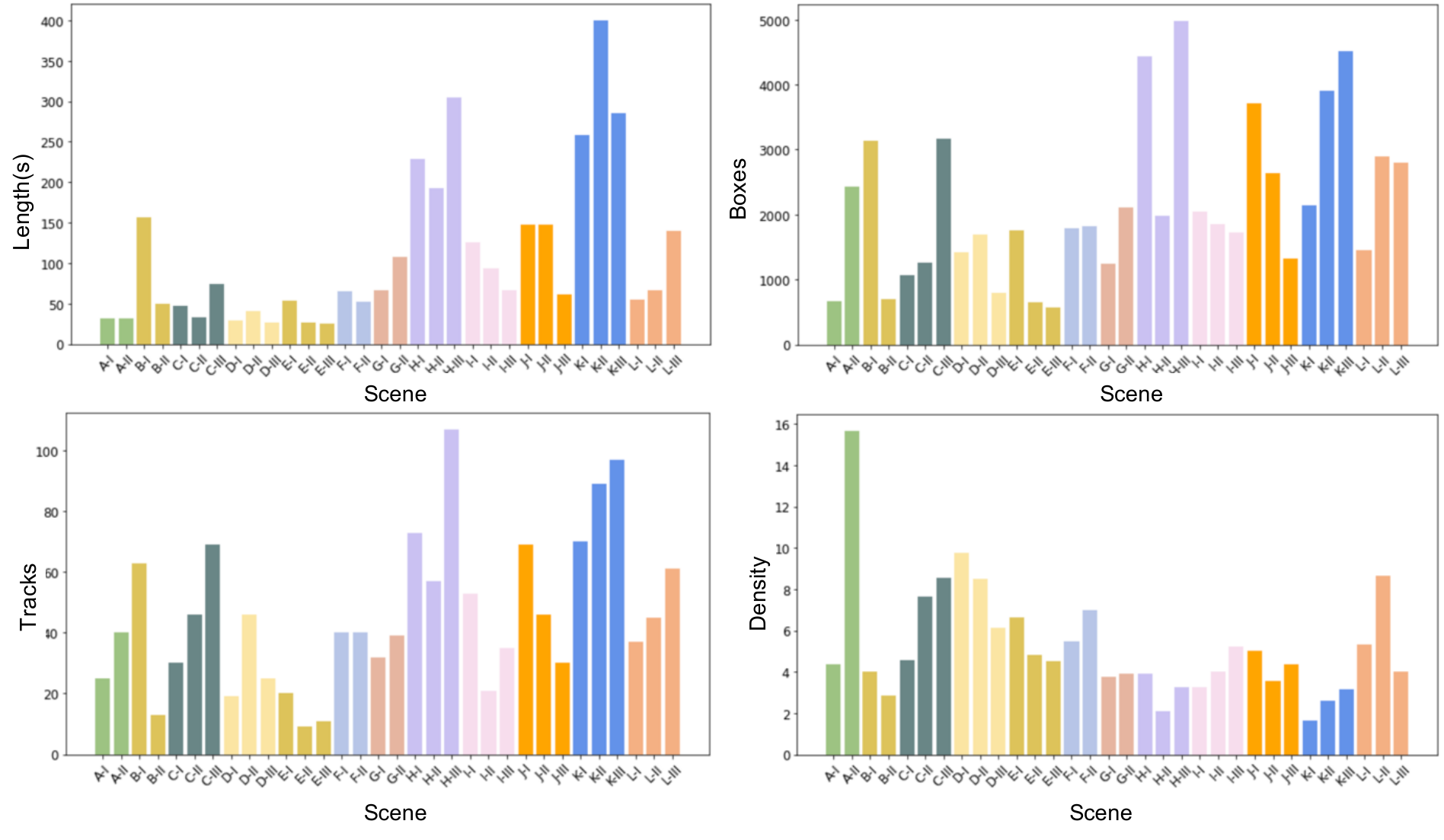}
    \caption{Overview of the proposed MMCT dataset. Different colors represent different scenes. There are 12 scenes in total, and each scene includes two or three sequences.}
  \label{Details of the dataset}
\end{figure}

The statistical information about these scenes is exhibited in Figure \ref{Details of the dataset}. There are in total of 32 sequences. For example, scene A includes two sequences of A-I and A-II. The lengths of the video for the two sequences are both 31 seconds. Sequence A-I contains 25 tracks with 673 bounding boxes, while A-II includes 40 tracks with 2,428 boxes. The column of Density represents the average number of pedestrians per image. Several real scenarios are shown in Figure \ref{data_scene}, where the FoVs of each sequence may be spatially overlapping or non-overlapping at different timestamps. Besides, it can be observed that different sequences captured by different devices in the same scene exhibit variations in color and brightness. This poses challenges for cross-camera data association.

Table \ref{Comparison of MTMMC dataset} compares our MMCT dataset to two other multi-moving-camera pedestrian tracking datasets. Similar to MMCT, the video sequences from driving recorders provided in \cite{lee2015road} (abbreviated as UWDR in this paper) and DHU-MTMMC focus on on-road pedestrian tracking. As shown, MMCT holds a higher resolution and includes the largest number of tracks, boxes, and sequences. Besides, the longer sequence length makes MMCT more appropriate for long-term on-road pedestrian tracking applications compared to the two other datasets.


\begin{table}[]
\caption{Comparison of different pedestrian tracking datasets collected under multiple moving cameras. UWDR refers to the dataset proposed in \cite{lee2015road}.}
\begin{tabular}{cccc}
\hline
Information & UWDR \cite{lee2015road}  & DHU-MTMMC \cite{zhang2021pedestrian}    & MMCT (ours)    \\ \hline
Cams        & 4                    & 3            & 3              \\
Scenes        & 3                    & 6            & 12              \\
Sequences   & 6                    & 14           & 32             \\
Tracks      & 126         & 180 & 1,457 \\
Boxes       & -                    & 8,371        & 68,738         \\
Length (s)  & 341                  & 422          & 3,385          \\
Resolution  & 1280 $\times$ 720             & 1920 $\times$ 1080    & 1920 $\times$ 1080      \\
Weather     & Sunny                & Sunny        & Cloudy         \\
Year        & 2015                 & 2021         & 2023           \\ \hline
\end{tabular}
\label{Comparison of MTMMC dataset}
\end{table}

\subsection{DHU-MTMMC dataset}
DHU-MTMMC dataset has 14 sequences distributed in 6 scenes with each scene covering the interaction from two or three moving cameras on different cars. The dataset is collected under sunny weather conditions, and the camera motion is relatively smooth. The total length of all sequences is 422 seconds with the shortest length being 9 seconds and the longest being 86 seconds. The highest pedestrian density of all sequences is 9.91, and the lowest density of those is 0.98. Both DHU-MTMMC and MMCT datasets hold the same image resolution of 1920 $\times$ 1080 pixels, however, the latter one is much larger in terms of the dataset scale and more complex concerning the driving scenarios.


\subsection{MOT17 Dateset}
MOT17 dataset \cite{milan2016mot16} is a widely used pedestrian tracking dataset for MOT tasks. It is comprised of 7 training video sequences and 7 testing video sequences. It only involves a single camera in each scenario. The number of tracks is 1,331 and the number of total frames is 11,235.

\subsection{DukeMTMC-reID Dateset}
DukeMTMC-reID is an extension of the DukeMTMC tracking dataset \cite{ristani2016performance} for person re-identification. There are 1,404 identities appearing in over two cameras and 408 identities (distractor ID) who appear in only one camera. To be specific, 702 IDs are selected as the training set, and the left IDs are used as the testing set.

\section{Experiments}

\subsection{Implementation Details}
We have implemented several SCT trackers, and most settings are kept the same as the original paper. FairMOT has shown a satisfactory result, hence we utilize the pre-trained model from \cite{zhang2021fairmot} for detection and embedding in SCT. The DLA-34 is employed as the backbone network to extract high-quality features. Several datasets, including the ETH, CityPerson, CalTech, MOT17, CUHK-SYSU, and PRW, are used to train the detection branch or Re-ID branch. The details can be found in \cite{zhang2021fairmot}. The large-scale training sets guarantee better detection and embedding results. 


We use MOT17 dataset as the training data for Linker. The positive samples consist of tracklets cut from annotated trajectories. Negative samples are made through ground-truth tracklets with random spatiotemporal noise. The ratio of positive and negative samples is set as $1 : 3$. Adam is used as the optimizer with an initial learning rate of 0.001. We train it for 60 epochs with a cosine annealing learning rate schedule. In training, we use the fully connected network to predict the score with smooth cross-entropy \cite{szegedy2016rethinking} loss as the objective function, which denotes if two tracklets belong to the same ID. In inference, a Softmax function layer is added after the MLP to output the score.

During inference, we use tracklets generated from the SCT tracker as the input of Linker. $N$ in Section III-B is set to be 30. Thus, the input tracklet has a size of $30 \times 5$. The temporal threshold $\sigma _{t}$ and spatial threshold $\sigma _{s}$ are set as $\left ( 0,10 \right )$ and $90$, respectively. The association confidence score threshold $\sigma _{a}$ is set as 0.5.

In the multi-camera trajectory association process, the cross-camera association threshold $\alpha$ is defined as 0.5 for the MMCT dataset and 0.8 for the DHU-MTMMC dataset to ensure optimal association performance. In the ablation experiments on scene style transfer, for each scene, the color distribution of the images captured by Device 1 is used as a reference for color, and style transfer is applied to the images from other cameras accordingly. The original images from Device 1 and the color-transferred images from other devices are then used as inputs to the MTMMC tracking pipeline. The threshold $\alpha$ remains the same as in the inter-camera tracking.

\subsection{Evaluation Metrics}
We follow the standard SCT and MTMC evaluation protocols to evaluate our method, where the standard ID score metrics \cite{ristani2016performance} and CLEAR MOT metrics \cite{bernardin2008evaluating} are reported. ID score metrics calculate the trajectory level ID precision (IDP), ID recall (IDR), and identity F1 score (IDF1). CLEAR MOT metrics include multi-object tracking accuracy (MOTA), identity switches (IDS), mostly tracked targets (MT), mostly lost targets (ML), and fragments (FRAG). We also report higher order metric for evaluating multi-object tracking (HOTA) \cite{luiten2021hota}, which is a new tracking metric that explicitly balances the effect of performing accurate detection, association, and localization. Of all the metrics above, IDF1 and HOTA are used to evaluate different aspects of tracking performance. MOTA is computed based on False Positive (FP), False Negative (FN), and IDS. In general, the numbers of FP and FN are usually larger than IDS, therefore MOTA tends to focus on the detection performance. IDF1 evaluates the identity preservation ability and focuses more on the association performance. 





\subsection{Results of Single Camera Tracking with Linker}

\begin{table*}[]
\caption{Single camera tracking results on MMCT dataset. For each pair of comparisons, the first row presents the original tracker; the second, third, and fourth rows show the corresponding tracker with AFLink \cite{du2022strongsort}, TransLink \cite{zhang2023translink}, and the proposed Linker applied, respectively. The number in ``()” besides the IDF1 score is the improvement after Linker is applied. The two best results for IDF1 are bolded and highlighted in red and blue. The best results for other metrics are bolded in black.}
\centering
\begin{tabular}{cccccccc}
\hline
Method             & IDF1 (\%) $\uparrow$     & MOTA (\%) $\uparrow$     & HOTA (\%) $\uparrow$     & MT (\%) $\uparrow$      & ML (\%) $\downarrow$      & IDS $\downarrow$           & FRAG $\downarrow$         \\
\hline
DeepSORT \cite{wojke2017simple} & 27.7 & 20.0 & 36.8 & \textbf{566} & 311 & 17,704 & \textbf{3,505} \\
DeepSORT+AFLink & 31.4 & 20.3 & 39.7 & 565 & 311 & 17,478 & 3,529 \\
DeepSORT+TransLink & \textcolor{red}{\textbf{33.9}} & \textbf{28.2} & \textbf{41.1} & 449 & 334 & \textbf{11,227} & 4,771  \\
DeepSORT+Linker & \textcolor{blue}{\textbf{32.4} (+4.7)}  & 20.7 & 40.2 & 563 & \textbf{311} & 17,254 & 3,552 
\\ \hline
Tracktor \cite{bergmann2019tracking}               & 47.1          & 37.2          & 53.2          & \textbf{538}          & \textbf{314}          & 6,535         & \textbf{3,676} \\
Tracktor+AFLink & 49.0 & 37.5 & 53.9 & 538 & 315 & 6,341 & 3,689 \\
Tracktor+TransLink & \textcolor{red}{\textbf{50.0}} & \textbf{40.1} & \textbf{54.3} & 483 & 333 & \textbf{4,516} & 4,054 \\
Tracktor+Linker & \textcolor{blue}{\textbf{49.5} (+2.4)} & 37.7 & 54.2 & 537 & 315 & 6,198 & 3,704  
\\ \hline
JDE \cite{wang2020towards}& 46.8 & 34.2 & 56.9 & \textbf{507} & \textbf{350} & 5,543 & \textbf{4,321} \\
JDE+AFLink & 49.4 & 34.6 & 58.8 & 507 & 350 & 5,285 & 4,329 \\
JDE+TransLink & \textcolor{red}{\textbf{50.7}} & \textbf{36.4} & \textbf{59.1} & 458 & 363 & \textbf{3,791} & 4,607 \\
JDE+Linker & \textcolor{blue}{\textbf{49.7} (+2.9)} & 34.8 & 58.9 & 505 & 351 & 5,137 & 4,331 
          \\ \hline
TrackFormer \cite{meinhardt2021trackformer}        & 47.5          & 45.2          & 55.1          & \textbf{629} & \textbf{275}          & 3,399          & \textbf{3,242} \\
TrackFormer+AFLink & \textcolor{blue}{\textbf{52.8}}          & 45.8          & 57.9          & 629          & 275          & 3,027          & 3,246          \\
TrackFormer+TransLink & 49.1 & 45.6 & 55.9 & 616 & 277 & 3,157 & 3,336 \\
TrackFormer+Linker & \textcolor{red}{\textbf{54.0}} \textcolor{red}{(+6.5)} & \textbf{46.2} & \textbf{58.7} & 625          & 276          & \textbf{2,817} & 3,252  
\\ \hline
FairMOT \cite{zhang2021fairmot}           & 58.3          & 49.2          & 61.0          & \textbf{652} & 252          & 2,642          & \textbf{3,195} \\
FairMOT+AFLink     & \textcolor{blue}{\textbf{59.5}}          & 49.5          & 61.0          & 652          & 252          & 2,473          & 3,199          \\
FairMOT+TransLink & 58.4 & 49.4 & 60.7 & 646 & 256 & 2,481 & 3,235 \\
FairMOT+Linker     & \textcolor{red}{\textbf{59.9}} \textcolor{red}{(+1.6)} & \textbf{49.6} & \textbf{61.1} & 651          & \textbf{252} & \textbf{2,405} & 3,198 
\\ \hline

\end{tabular}
  \label{comparison of different methods on linker}
\end{table*}

\begin{table}[]
\caption{Comparison of AFlink, TransLink, and Linker in terms of size and running time in FairMOT on MMCT and DHU-MTMMC datasets.}
\centering
\begin{tabular}{cccc}
\hline
\multirow{2}{*}{Method}   & \multirow{2}{*}{Size (M)}     & \multicolumn{2}{c}{Time (s)} \\
          &              & MMCT   & DHU-MTMMC  \\ \hline
AFLink    & 4.2          & 14.3          & 2.6          \\
TransLink & 23           & 75.9          & 5.2          \\
Linker    & \textbf{2.7} & \textbf{9.6}  & \textbf{1.7} \\ \hline
\end{tabular}
\label{Comparison of model size and running time}
\end{table}

A better SCT method usually contributes to a better performance of the multi-camera-tracking framework. 
In this paper, we provide the tracking result of five SCT methods, \textit{i.e.}, DeepSORT \cite{wojke2017simple}, Tracktor \cite{bergmann2019tracking}, JDE \cite{wang2020towards}, TrackFormer \cite{meinhardt2021trackformer}, and FairMOT \cite{zhang2021fairmot}. To be specific, DeepSORT learns a deep association metric and establishes measurement-to-track associations using nearest neighbor queries in visual appearance space. Tracktor tackles multi-object tracking by exploiting the regression head of a detector to perform temporal realignment of object bounding boxes. JDE characterizes a simple and fast association method that works in conjunction with the joint mode and reports the first (near) real-time MOT system. Trackformer is an end-to-end MOT approach based on an encoder-decoder Transformer architecture. It represents a new tracking-by-attention paradigm with autoregressive track queries and obtains competitive results. 

To explicitly see how Linker performs, we also implement the AFlink proposed in StrongSORT \cite{du2022strongsort} and TransLink \cite{zhang2023translink} for a comparison. AFLink only takes the (frame index, x, y) as input and adopts the serial operations of the temporal module and fusion module, while our proposed Linker takes two more values of (w, h), indicating the box size, as inputs and utilizes a parallel structure in the spatiotemporal module. Resorting to a better feature representation, TransLink incorporates self-attention module with both motion and appearance features as the input, which increases the size of the model to some extent.



\textbf{Quantitative results on MMCT.} The experimental results for SCT on MMCT dataset are reported in Table \ref{comparison of different methods on linker}. For each pair of comparisons, the original method is presented in the first row, and the method enhanced with AFLink, TransLink, and the proposed Linker is presented in the second, third, and fourth rows, respectively. As presented, FairMOT outperforms other trackers by a large margin. Similar to FairMOT, both JDE and DeepSORT are real-time MOT systems, but their performances fall far behind FairMOT. TrackFormer performs slightly better than JDE, while also cannot come close to FairMOT. FairMOT utilizes CenterNet to obtain the bounding box and then uses multi-layer fusion to generate high-quality embedding appearance features from the center of the target, which clearly helps to achieve the best results.


Moreover, as shown in Table \ref{comparison of different methods on linker}, AFLink, TransLink, and Linker all enhance the tracking performances. To be specific, Linker brings larger improvement over the AFLink in all five trackers. TransLink outperforms Linker in the first three trackers, while Linker exhibits better improvements in TrackFormer and FairMOT by another 4.9\% and 1.5\% in IDF1 than Linker. Linker performs better than AFLink, owing to more box information utilized and a more effective spatio-temporal module employed. 
There are significant improvements on the IDF1 score with Linker applied, ranging from +1.6\% to +6.5\%, since Linker aims at tackling missing associations. Besides, the application of Linker brings different levels of performance improvement for different methods. It can be concluded that the increase in IDF1 score is larger in terms of poorer trackers due to more missing associations. Specifically, the IDF1 score of TrackFormer increases by 6.5\% with the Linker added. The MOTA metrics of each method with Linker applied roughly remain the same compared to the original method since we do not focus on addressing the detection part. HOTA, which is associated with both the detection and association part, is enhanced as well. 

To make a more comprehensive comparison among AFLink, TransLink, and the proposed Linker, we also give the model size and test the running time of the three methods based on the tracking results of FairMOT on all video sequences of MMCT and DHU-MTMMC datasets, respectively, as shown in Table \ref{Comparison of model size and running time}. We observe that Linker has the smallest model size and takes the shortest running time with competitive tracking results achieved which can be seen in Table \ref{comparison of different methods on linker} and Table \ref{single comparison}. To be mentioned, Linker strikes a trade-off between time and performance, which is the top priority for some applications that require rapid deployment in real life.

\begin{figure*}[h]
  \centering
  \includegraphics[width=\linewidth]{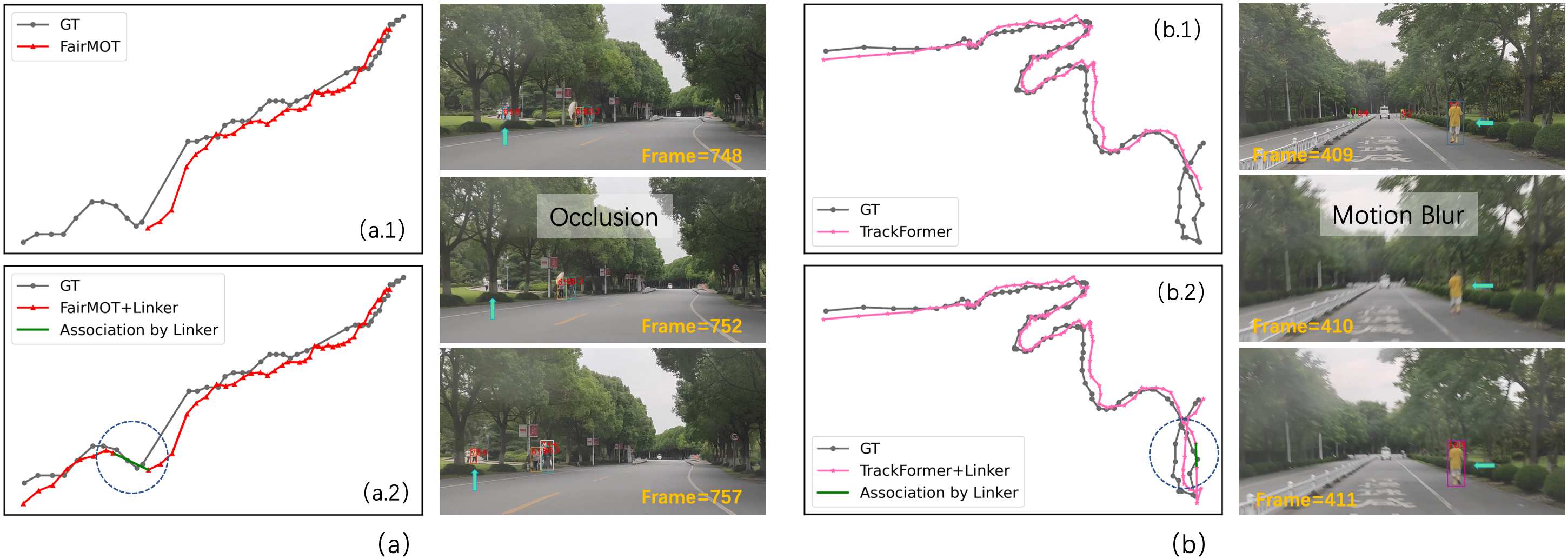}
  \caption{The exemplary cases in MMCT that Linker takes effects to solve the ID switches. The Linker can cooperate with different trackers: (a) FairMOT, the presented trajectories are sampled at selected continuous 41 frames from sequence K-II; (b) TrackFormer, the presented trajectories are sampled at selected continuous 100 frames from sequence K-II.  The green line inside the blue dash circle indicates that the ID switch caused by either occlusion or motion blur is fixed by Linker.}
  \label{sinle_camera_trajectory}
\end{figure*}

\textbf{Qualitative results on MMCT.} To better visualize the improvement with Linker, we show some qualitative comparison results on the sequence K-II in Figure \ref{sinle_camera_trajectory}. Each color of the bounding box signifies a distinct predicted object ID. Figure \ref{sinle_camera_trajectory}(a) shows the tracking results of FairMOT. In this scene, a pedestrian with the pointed arrow is completely occluded for several frames and then reappears from occlusion with another ID assigned at frame 757. Thus the length of the trajectory is shorter than the GT trajectory, as is shown in (a.1). Figure \ref{sinle_camera_trajectory}(b) shows a tracking example from TrackFormer. A pedestrian in yellow with the pointed arrow changes ID at frame 411 due to motion blur at frame 410, which leads to blurred vision. ID switch happens in both (a.1) and (b.1) due to occlusion and motion blur, respectively. MMCT dataset is collected under multiple moving cameras, in which case motion blur and occlusion exist extensively in the sequences. ID switches are observed frequently in a pure SCT tracker. With the proposed Linker applied, the missing association errors can be effectively fixed, as exhibited inside the blue circle in (a.2) and (b.2), respectively. The green line indicates two separate short tracklets are associated together by Linker.

\textbf{Results on DHU-MTMMC.}  We also conduct comparative experiments on DHU-MTMMC dataset. The competing methods are the same as those used in MMCT dataset. The results are reported in Table \ref{single comparison}. As presented, the results of FairMOT also rank first, and the application of Linker in all trackers helps yield better performance. Furthermore, Linker outperforms TransLink and AFLink in most trackers regarding tracking performance.

\begin{table}[]
\caption{Results of different single-camera tracking methods with AFLink, TransLink, and the proposed Linker on DHU-MTMMC dataset. The two best results for IDF1 are bolded and highlighted in red and blue.}
\centering
\begin{tabular}{cccc}
\hline
Method               & IDF1 (\%) $\uparrow$    & MOTA (\%) $\uparrow$    & IDS $\downarrow$         \\ \hline
DeepSORT \cite{wojke2017simple}            & 33.0          & 25.0          & 1,861         \\
DeepSORT+AFLink & \textcolor{blue}{\textbf{39.5}}  & 25.6 & 1,808 \\ 
DeepSORT+TransLink & \textcolor{red}{\textbf{\textbf{42.1}}} &  \textbf{34.1} & \textbf{988} \\
DeepSORT+Linker & 39.1 (+6.1)  &25.7 & 1,796 
\\ \hline
Tracktor \cite{bergmann2019tracking}            & 54.2         & 39.7         & 594  \\
Tracktor+AFLink & 55.4 & 40.0 & 572 \\
Tracktor+TransLink & \textcolor{red}{\textbf{\textbf{56.3}}} & \textbf{42.1} & \textbf{416} \\
Tracktor+Linker & \textcolor{blue}{\textbf{55.6} (+1.4)}  & 40.2 & 556 
\\ \hline
JDE \cite{wang2020towards}           & 66.5          & 62.3          & 434          \\
JDE+AFLink & \textcolor{blue}{\textbf{69.4}} & 62.6 & 412 \\
JDE+TransLink & 69.2 & \textbf{63.0} & \textbf{298} \\
JDE+Linker & \textcolor{red}{\textbf{69.7}} \textcolor{red}{(+3.2)} & 62.7 & 403 
\\ \hline
TrackFormer \cite{meinhardt2021trackformer}         & 62.1          & 62.0          & 300  \\
TrackFormer+AFLink & 64.9 & 62.5 & 260 \\
TrackFormer+TransLink & \textcolor{blue}{\textbf{65.4}} & 62.7 & 251 \\
TrackFormer+Linker & \textcolor{red}{\textbf{67.4}} \textcolor{red}{(+5.3)}  & \textbf{63.0} & \textbf{237} 
\\ \hline
FairMOT \cite{zhang2021fairmot}       & 70.6          & 63.4          & 279          \\
FairMOT+AFLink & 70.9 & 63.5 & 273 \\ 
FairMOT+TransLink & \textcolor{blue}{\textbf{71.0}} & 63.5 & \textbf{269} \\
FairMOT+Linker & \textcolor{red}{\textbf{71.4}} \textcolor{red}{(+0.8)}  & \textbf{63.6} & 270 
\\ \hline
\end{tabular}
\label{single comparison}
\end{table}

\subsection{Results of Multi-camera Tracking}
\begin{figure}[h]
  \centering
  \includegraphics[width=0.9\linewidth]{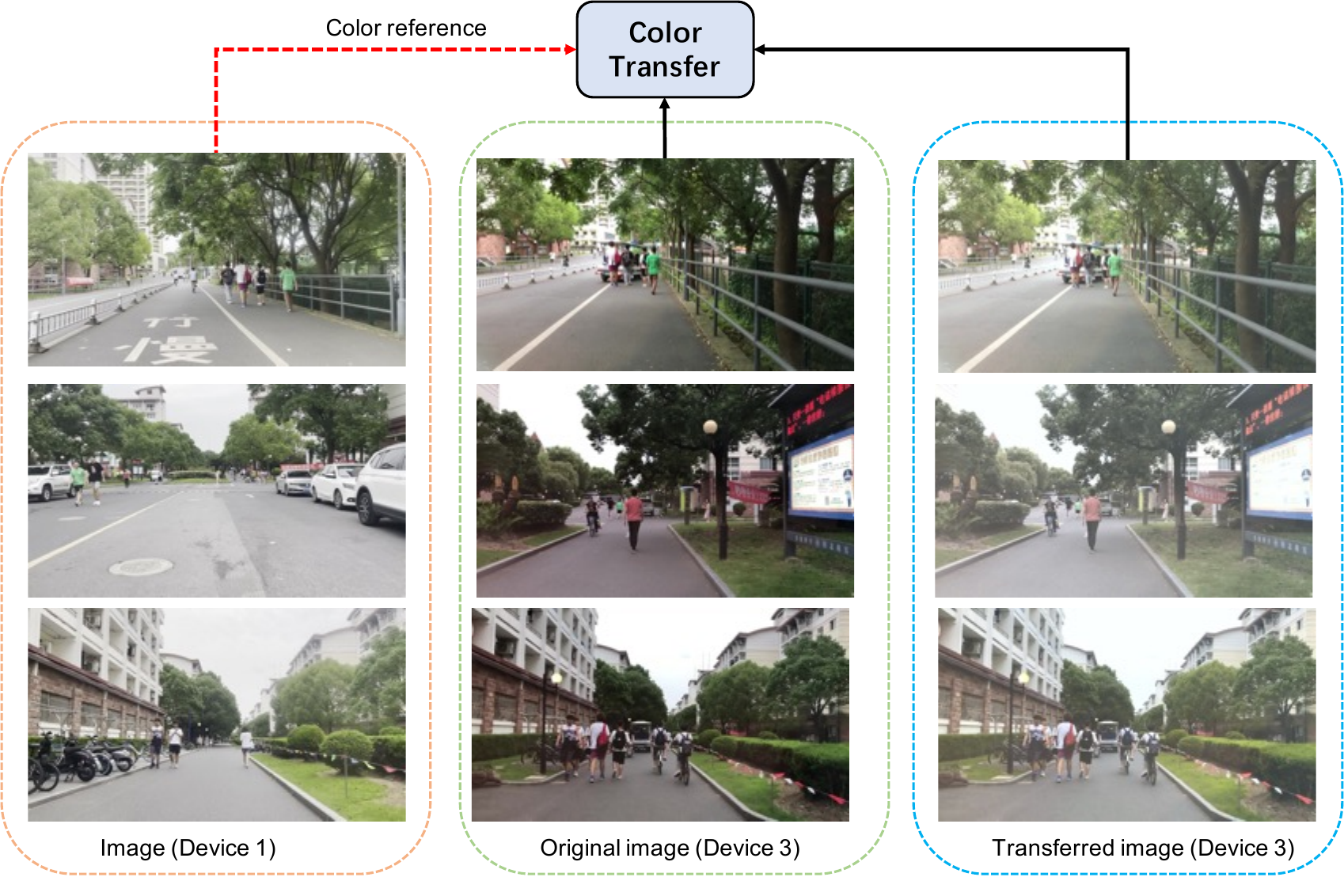}
  \caption{Illustration of scene style transfer. The images captured by Device 1 in each scene are used as style reference images for the images from other cameras, and the style transfer module is applied to transform the images from other cameras.}
  \label{visualization of color transfer}
\end{figure}

\textbf{Ablation study of color transfer module.} Tracking across multiple moving cameras is quite challenging. Due to the variations in illumination, camera settings, \textit{etc.}, the appearance features of the same person in different cameras may differ greatly. Therefore, the color transfer module is utilized to mitigate the impact. To test the significant impact of style transfer on cross-camera tracking, ablation experiments are conducted, as shown in Table \ref{color_transfer}. 
``w/o.'' indicates the absence of the style transfer module, where the original images from each camera are used as inputs. ``w/.'' indicates the inclusion of the style transfer module, where the original image from Device 1 and the color-transferred images from other cameras as inputs. In both MMCT and DHU-MTMMC datasets, there are noticeable style differences among images captured by different cameras in the same scene. Examples of image style transfer can be seen in Figure \ref{visualization of color transfer}. The experimental results demonstrate that the application of the style transfer module leads to an increase in IDF1 values by 0.6\% and 3.2\% on the MMCT and DHU-MTMMC datasets, respectively. This indicates that incorporating style transfer is beneficial for extracting cross-camera consistent features, thereby reducing the feature variations between trajectories of the same person captured by different cameras.

Moreover, as shown in Table \ref{color_transfer}, the results of all metrics on DHU-MTMMC are much higher than those on MMCT. As mentioned in Section IV-A, MMCT dataset is more challenging, collected under a more complicated environment, such as cloudy weather, various driving routes with frequent camera shakes, \textit{etc.} Besides, the sequences are longer with long-term occlusions observed, which requires more novel and effective methods to obtain a promising result.

\textbf{Effect of different association thresholds.} We also conduct experiments related to threshold $\alpha$, as used in the Jonker-Volgenant algorithm \cite{jonker1987shortest} for the distance matrix $D$ (see Eq.(\ref{D})), in the multi-camera data association stage on MMCT and DHU-MTMMC datasets, respectively. Specifically, we vary $\alpha$ from 0.3 to 0.6 for the MMCT dataset and vary $\alpha$ from 0.7 to 1.0 on DHU-MTMMC. As shown in Table \ref{effect of threshold}, the best result is achieved when $\alpha$ is set as 0.5 for MMCT and 0.8 for DHU-MTMMC, and no further improvement with the increase of $\alpha$. Therefore, it is necessary to select a proper association threshold to filter out the mismatches.

\begin{table}[]
\caption{The ablation study of the color transfer module. CT refers to color transfer.}
\centering
\begin{tabular}{ccccc}
\hline
Dataset                    & CT & IDF1 (\%)            & IDP (\%)      & IDR (\%)      \\ \hline
\multirow{2}{*}{MMCT}      & w/o.           & 48.4                 & 51.0          & 43.7          \\
                           & w/.            & \textbf{49.0 (+0.6)} & \textbf{52.1} & \textbf{43.8} \\ \hline
\multirow{2}{*}{DHU-MTMMC} & w/o.           & 55.1                 & 59.9          & 50.8          \\
                           & w/.            & \textbf{58.3 (+3.2)} & \textbf{64.6} & \textbf{52.9} \\ \hline
\end{tabular}
\label{color_transfer}
\end{table}

\begin{table}[]
\caption{Influences from the association threshold $\alpha$ in multi-camera tracking on MMCT and DHU-MTMMC datasets.}
\centering
\begin{tabular}{ccccc}
\hline
Dataset                              & $\alpha$          & IDF1 (\%)                      & IDP (\%)                       & IDR (\%)                       \\ \hline
\multirow{4}{*}{MMCT} & 0.3 & 48.1                           &50.6                           & 43.4                           \\
                                     & 0.4 & 48.3 & 50.9 & 43.7 \\
                                     & 0.5 & \textbf{48.4}                           & \textbf{51.0}                           & \textbf{43.7}                           \\
                                     & 0.6 & 48.3                           & 50.8                           & 43.6                           \\ \hline
\multirow{4}{*}{DHU-MTMMC} & 0.7 & 54.8                           & 59.5                           & 50.5                           \\
                                     & 0.8 & \textbf{55.1}                           & \textbf{59.9}                           & \textbf{50.8}                           \\
                                     & 0.9 & 54.1                          & 58.8                           & 49.9                          \\
                                     & 1.0 & 53.5                          & 58.2                           & 49.3                           \\ \hline
\end{tabular}
  \label{effect of threshold}
\end{table}

\begin{figure*}[h]
  \centering
  \includegraphics[width=0.95\linewidth]{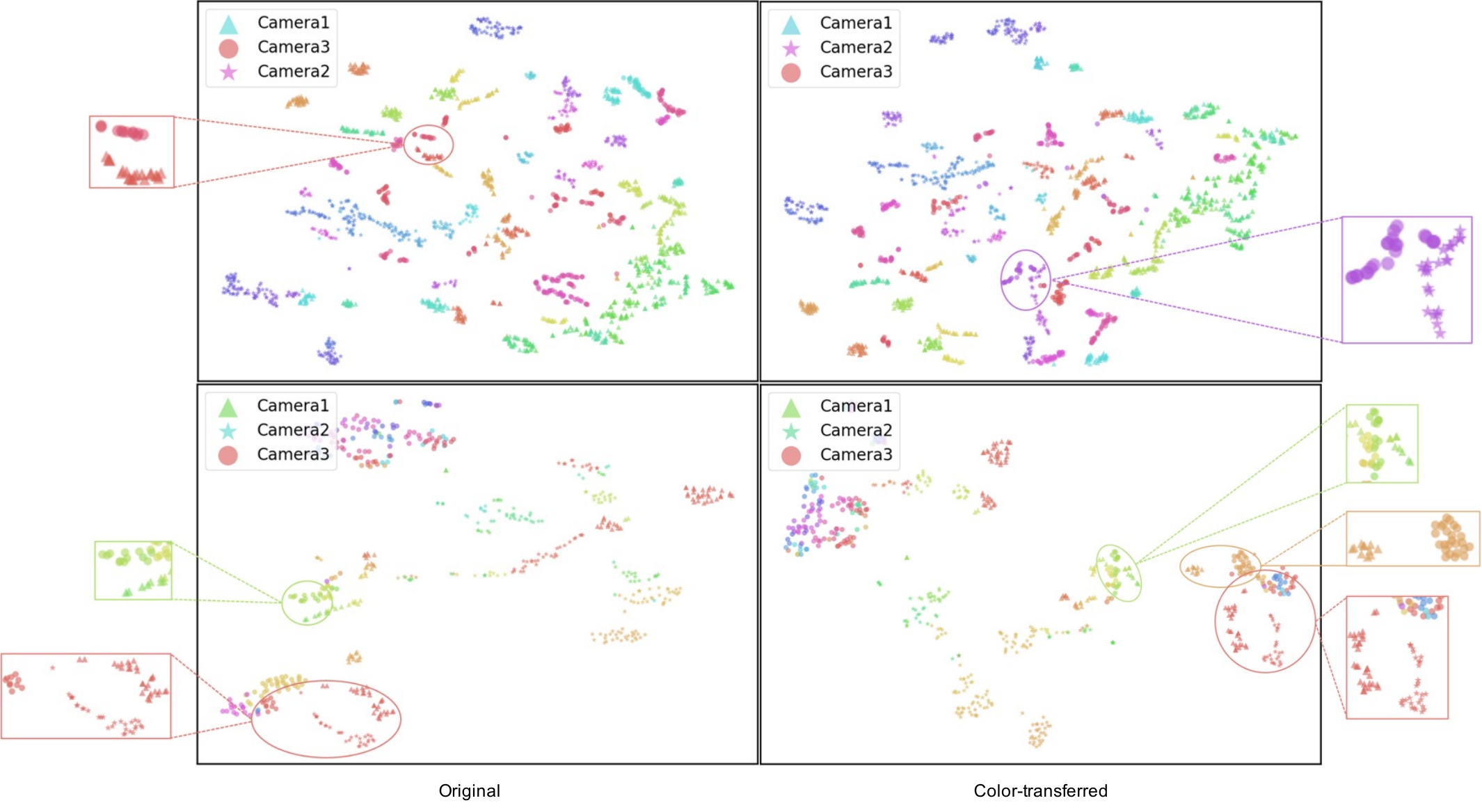}
  \caption{Visualization of feature embeddings of pedestrians from Scene J on MMCT in the first row and Scene F on DHU-MTMMC in the second row for different identities from multiple cameras using $t$-SNE. Different colors represent different tracklets, while different shapes represent different cameras. Embeddings from color-transferred images representing the same person in different cameras tend to be grouped closer than the original image. There are more clusters of feature embeddings of the same identity from multiple cameras for feature embeddings in the color-transferred images than in the original images. The leftmost and rightmost small boxes are the enlarged display of several clusters of feature embeddings.}
  \label{tsne_reid_fairmot_comparison}
\end{figure*}

\textbf{Visualization of feature embeddings.} We use $t$-Distributed Stochastic Neighbor Embedding ($t$-SNE) to show the Re-ID features in Figure \ref{tsne_reid_fairmot_comparison}. The visualization of Re-ID features of pedestrians from the original images is presented in the left column and the color-transferred images in the right column, respectively, from which we can clearly see that the features of the same identity from color-transferred images in different cameras tend to be grouped more compactly. Color transfer decreases the domain difference among images captured by different cameras. Hence, more robust embeddings can be obtained for the ICT data association.

\subsection{Discussion and Limitation}
The proposed MTMMC tracking baseline uses the embedding features extracted from the Re-ID branch of FairMOT for ICT owing to its efficiency and convenience. Another solution is to use the features typically for person re-identification tasks, which might result in a better performance. Moreover, the spatial information of the moving cameras is not fully utilized in the baseline model, which deserves more exploration.

\section{Conclusion}
Aiming at bringing together the vision information from different on-road vehicles, in this paper, we collect a dataset, called MMCT. To our best knowledge, this dataset is the largest one featuring multi-moving-camera tracking beyond real-world vehicle visions, which can provide a data basis to evaluate the MTMMC algorithms. Moreover, we propose an effective workflow for multi-pedestrian tracking across multiple moving cameras. The experimental results have demonstrated that the proposed appearance-free Linker model can boost SCT performances regardless of tracker types. It is lightweight and can be applied to any tracker to solve identity switch issues to some extent. To extract consistent feature embeddings across different cameras, we have utilized a color transfer method. Note that, any newer SCT tracker can be applied in this framework. In the future, we will take the spatial motion information of different moving vehicles into account to enhance the tracking efficiency across cameras.




\ifCLASSOPTIONcaptionsoff
  \newpage
\fi



\bibliographystyle{IEEEtran}
\bibliography{ieee}

%



%

\begin{IEEEbiography}[{\includegraphics[width=1in,height=1.25in,clip,keepaspectratio]{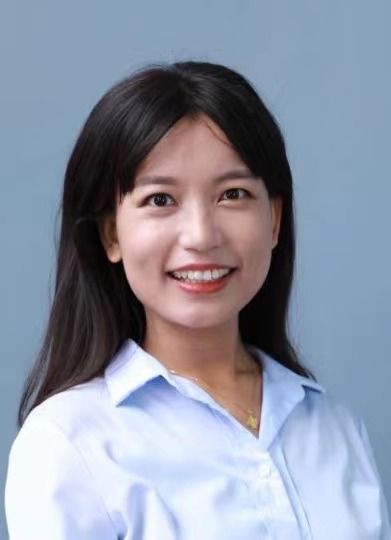}}]{Yanting Zhang}(Member, IEEE)
received the B.E. degree and Ph.D. degree in the School of Information and Communication Engineering from Beijing
University of Posts and Telecommunications in
2015 and 2020, respectively. She used to be a visiting scholar at the University of Washington (Seattle) from 2018 to 2019. She is currently an Assistant Professor in the School of Computer Science and Technology at Donghua University. Her research interests include computer vision and video/image processing.
\end{IEEEbiography}

\begin{IEEEbiography}[{\includegraphics[width=1in,height=1.25in,clip,keepaspectratio]{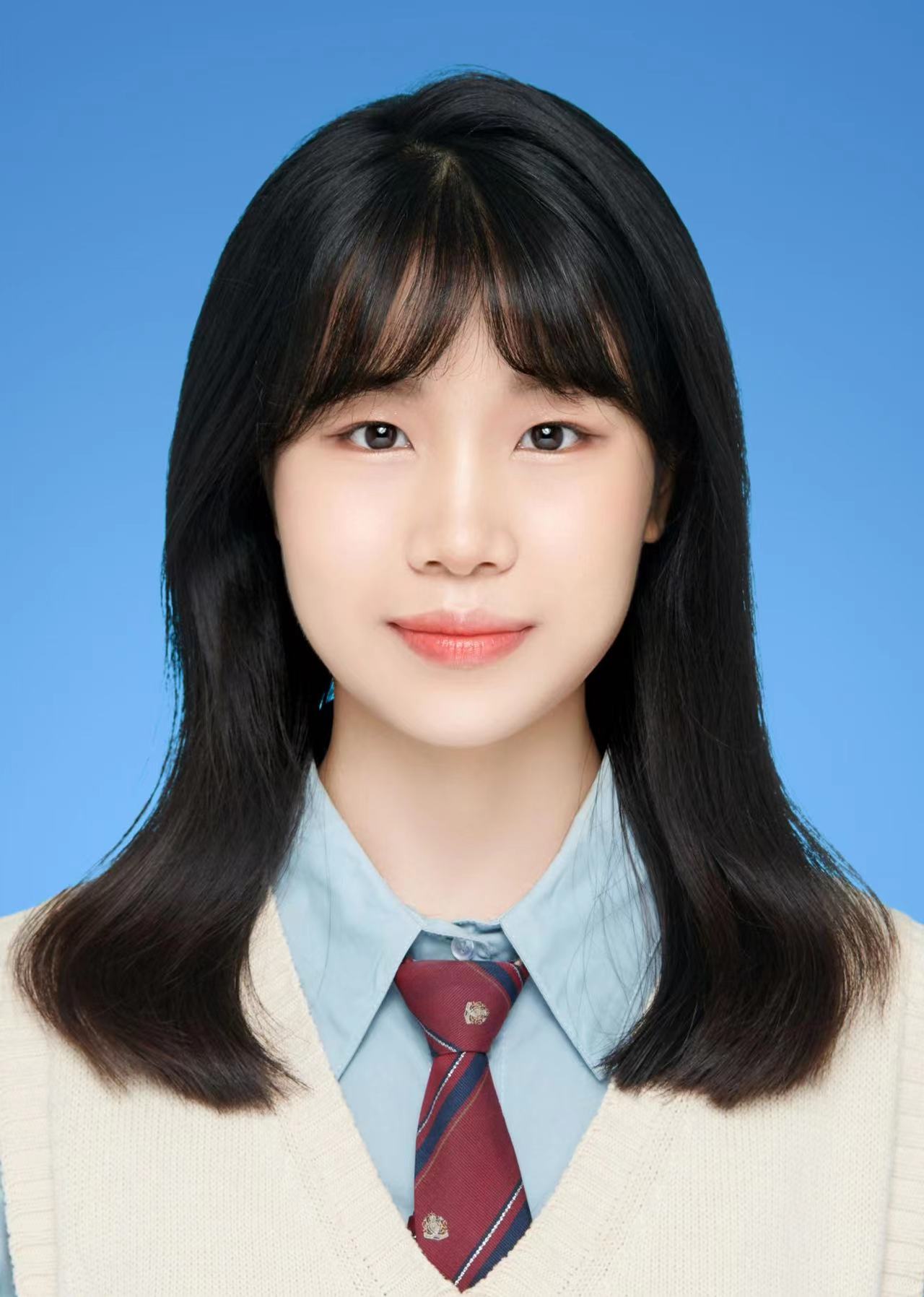}}]{Shuanghong Wang}(Graduate Student Member, IEEE) received the B.E. degree from Hubei University of Technology in 2021. She is currently working toward the M.S. degree in Donghua University. Her current research interests include computer vision and video/image processing.
\end{IEEEbiography}


\begin{IEEEbiography}[{\includegraphics[width=1in,height=1.25in,clip,keepaspectratio]{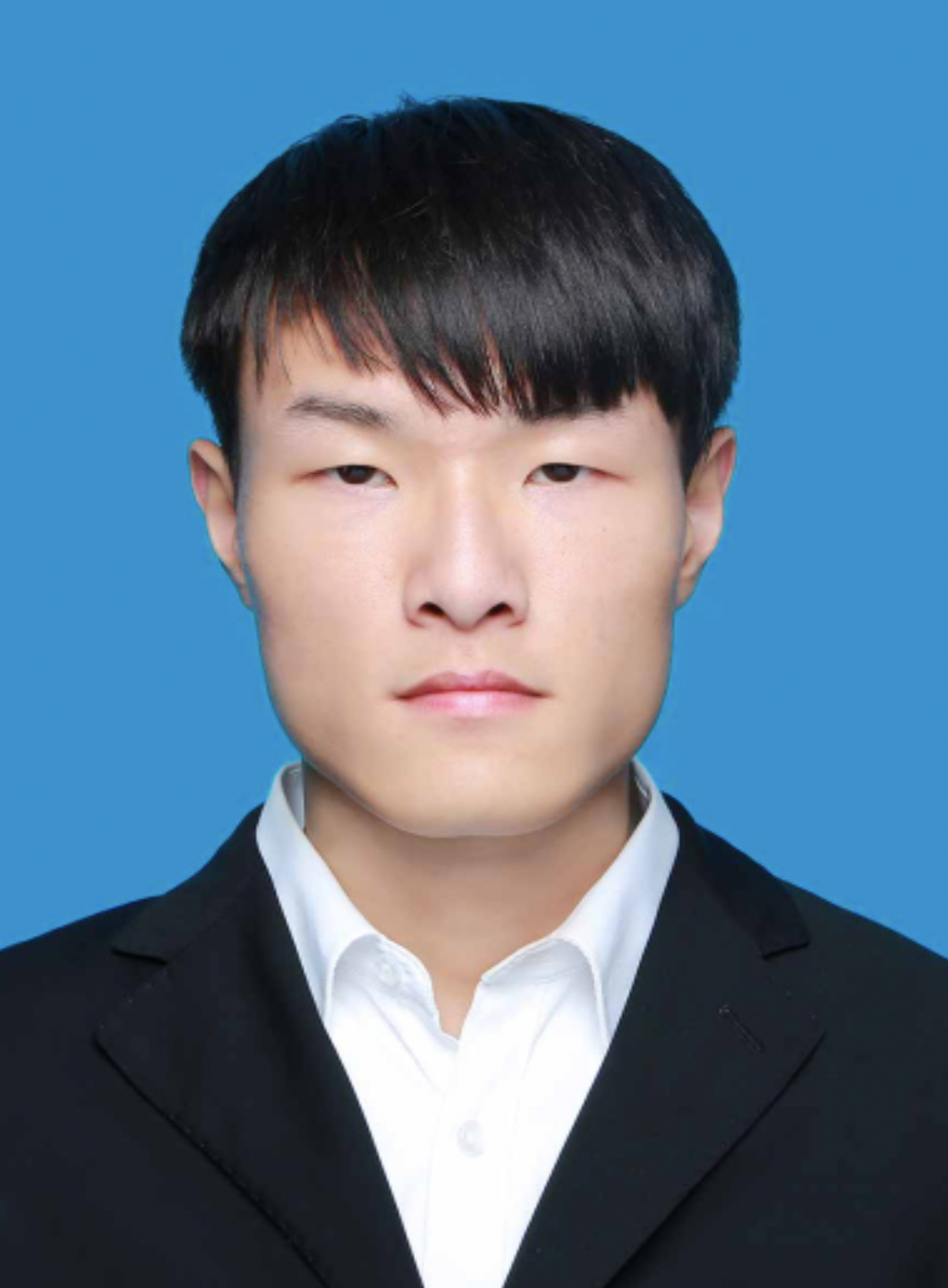}}]{Qingxiang Wang} received the B.E. degree from Donghua University in 2021. He is currently a master student in Hosei University. His research interests include computer vision and video/image processing.
\end{IEEEbiography}

\begin{IEEEbiography}[{\includegraphics[width=1in,height=1.25in,clip,keepaspectratio]{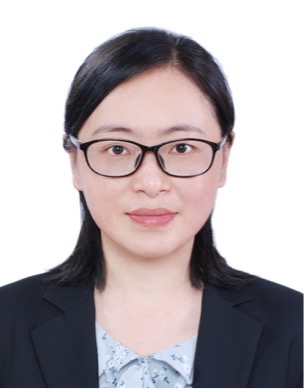}}]{Cairong Yan} received the Ph.D. degree in computer science and technology from Xi’an Jiaotong University, Shaanxi, China, in 2006. She is currently an associate professor and a master tutor. Her research interests include data mining, machine learning and big data processing.
\end{IEEEbiography}

\begin{IEEEbiography}[{\includegraphics[width=1in,clip]{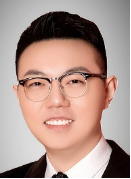}}]{Rui Fan}(Senior Member, IEEE)  
received the B.Eng. degree in automation from the Harbin Institute of Technology in 2015 and the Ph.D. degree in electrical and electronic engineering from the University of Bristol in 2018. He worked as a Research Associate at the Hong Kong University of Science and Technology from 2018 to 2020 and a Postdoc Fellow at the University of California San Diego between 2020 and 2021. He is currently a full Professor at Tongji University and Shanghai Research Institute for Intelligent Autonomous Systems. Rui was named in Stanford University List of Top 2\% Scientists Worldwide in 2022. His research interests include computer vision, deep learning, and robotics.
\end{IEEEbiography}




\end{document}